\documentclass[10pt]{article}

\usepackage[margin=1in]{geometry}
\usepackage{times}
\usepackage{microtype}
\usepackage{amsmath,amssymb,amsthm,mathtools}
\usepackage{booktabs}
\usepackage{tabularx}
\usepackage{graphicx}
\usepackage{xcolor}
\usepackage{enumitem}
\usepackage[numbers,sort&compress]{natbib}
\usepackage[colorlinks=true,allcolors=blue!55!black]{hyperref}
\usepackage[nameinlink,capitalize]{cleveref}
\usepackage{tikz}
\usetikzlibrary{arrows.meta,positioning,calc}

\newtheorem{proposition}{Proposition}
\newcommand{\doop}{\mathrm{do}}
\newcommand{\E}{\mathbb{E}}
\newcommand{\KL}{D_{\mathrm{KL}}}
\newcommand{\ObjSet}{\mathcal{V}}
\newcommand{\Cand}{\mathcal{C}}
\newcommand{\indep}{\perp\!\!\!\perp}
\newcommand{\method}{WhatIf}

\title{WhatIf: Counterfactual World Models over\\
Action-Grounded Causal Graphs for Robotic Manipulation}
\author{%
Xu Yang\textsuperscript{1}
\quad
Kailai Sun\textsuperscript{2,3,}\thanks{Corresponding authors: Kailai Sun and Qianchuan Zhao.}
\quad
Qianchuan Zhao\textsuperscript{1,}\footnotemark[1]
\\[0.8em]
\parbox{\dimexpr\textwidth-2\tabcolsep\relax}{\centering\small
\textsuperscript{1}Center for Intelligent and Networked Systems, Department of Automation, BNRist,\\
Tsinghua University, Beijing 100084, China\\[0.3em]
\textsuperscript{2}Singapore-MIT Alliance for Research and Technology Centre (SMART),\\
Massachusetts Institute of Technology, 138602, Singapore\\[0.3em]
\textsuperscript{3}Urban Mobility Lab, Massachusetts Institute of Technology,\\
Cambridge, MA 02139, United States}%
}
\date{}

\begin{document}
\maketitle

\begin{abstract}
World models predict the consequences of actions, but they are trained on factual transitions and represent the future of each action separately. Decisions, however, are comparisons: would another action have done better in the same physical situation? We introduce \method{}, a counterfactual world model that represents the joint law of all candidate futures of a manipulation scene by passing one abducted physical situation through an action-grounded causal graph. From a factual history, \method{} abducts hierarchical latent physics and transition noise; each candidate action generates a sparse object--factor graph whose topology and edge strengths specify the physical pathways it activates; and all candidate futures are rolled out under the shared situation. Shared type-conditioned functions instantiate the graph for any number of objects. Counterfactual predictions are anchored on observed outcomes, and the graph confines each action contrast to the objects that the action reaches. We show that factual data determine only the marginal future of each action, and hence the average action effect but neither the individual effect nor the probability that switching actions causes harm, and that anchoring with the average effect incurs an error equal to the conditional variance of the individual effect. A round-trip loss enforces action-invariant abduction without counterfactual labels. On the coupled futures we build Counterfactual Advantage Planning (CAP), which ranks plans by their advantage over an incumbent within shared situations, constrains the probability of harm, and converts each executed transition into hindsight advantages that update the physics posterior, the plan of the next attempt, and an amortized policy without a learned critic. Paired advantages reduce estimator variance by the factor $1-\rho$, where $\rho$ is the outcome correlation between branches, and cancel model errors common to both branches. We further derive when the observational transition $p(S_{t+1}\mid H_t,A_t)$ identifies $p(S_{t+1}\mid H_t,\doop(A_t))$ and where hidden action confounding or insufficient support breaks this equality. Paired simulator interventions evaluate counterfactual accuracy, graph recovery, compositional generalization, and decision quality.
\end{abstract}

\section{Introduction}

World models compress past observations into a state from which an agent predicts future trajectories and plans actions \citep{hafner2023dreamerv3,hansen2024tdmpc2}. Their success has largely been driven by expressive sequence models and predictive objectives. Given a history $H_t$ and the action taken by a behavior policy, a conventional world model learns
\begin{equation}
    p_{\theta}(S_{t+1}\mid H_t,A_t)
\end{equation}
by minimizing error on logged transitions. This objective does not require the model to separate stable physical mechanisms from regularities specific to the training policy, scene distribution, or object arrangement. A predictor may associate a visual configuration with its frequent successor without discovering which object caused which change. Consequently, world models often generalize poorly when masses, friction, object counts, layouts, or contact patterns differ from the training distribution. Scaling the predictor can improve interpolation, but it does not by itself make the learned transition invariant to such mechanism-preserving changes.

This limitation motivates a \emph{causal world model}: a model that represents the mechanisms by which objects and state factors influence one another, and whose predictions are composed from those mechanisms under intervention \citep{pearl2009causality,scholkopf2021causalrep}. In robotic manipulation, the distinction is concrete. A push creates a gripper--object contact, propagates force through selected object pairs, and leaves unrelated objects unchanged. If the same contact and force-transfer mechanisms are represented explicitly, they can be reused when the scene contains a different number or arrangement of objects. Causal structure is therefore not only an explanation layer; it is a hypothesis about which parts of the dynamics should remain stable across environments.

Mechanism discovery alone, however, does not answer the decision-time question that matters most: \emph{what would have happened in the same physical situation had the robot taken another action?} Logged data reveal only the outcome of the realized action. Predicting each candidate action independently can change latent context or stochastic realization together with the action, confounding action comparison with scenario variation. Counterfactual prediction instead follows abduction--intervention--prediction: infer the latent physical situation and exogenous realization responsible for the factual transition, replace only the action mechanism, and predict the resulting future \citep{pearl2009causality,buesing2019cfgps}. This distinction becomes decisive for control, because decisions are comparisons. Even a perfect conventional world model supplies only the marginal law of each action's future. These marginals fix the expected advantage of switching actions, but not how often the switch helps or harms in the same situation, nor which failures the action choice could have avoided. A \emph{counterfactual world model} must therefore represent the joint law of the futures of all candidate actions, coupled through one shared physical situation. This coupling yields lower-variance action comparisons, cancels model errors common to all candidates, bounds the probability of harm, and turns every executed transition into supervision for many actions that were never executed, without counterfactual $Q$-value labels.

\paragraph{Causal world models: progress and unresolved problem.}
Existing causal world models have pursued several complementary routes. Causal World Models infer object-level latent physical factors and use a structural model to predict outcomes under intervention \citep{li2020cwm}. CausalVAE as a plug-in learns a sparse directed acyclic graph over fixed latent dimensions, enforcing sparsity and acyclicity so that latent dependencies can support counterfactual reasoning \citep{ding2026causalvae}. DALI infers latent dynamics context from interaction histories to generalize to unseen physical conditions \citep{roder2025dali}. Object-centric models factor dynamics across entities \citep{biza2022factored,ferraro2023focus}; FIOC-WM, Dyn-O, and SPARTAN further expose modular interactions, separate dynamics-relevant factors, or infer sparse local graphs \citep{feng2025fioc,wang2025dyno,lei2025spartan}. These methods establish that physical context, object factorization, and local structure improve generalization. Nevertheless, they leave a common gap. A fixed latent DAG cannot naturally represent variable object sets or contacts created by a particular action; an unstructured context vector does not specify which edge it modulates; and a learned local graph is not necessarily regenerated under a replacement action while holding the factual physical realization fixed. Thus current causal world models do not yet provide an executable account of how an intervention changes both the topology and strength of a manipulation mechanism.

\paragraph{Counterfactual world models: progress and unresolved problem.}
Counterfactually-Guided Policy Search infers exogenous variables from an observed episode and evaluates alternative policies in the same inferred scenario \citep{buesing2019cfgps}. Work on confounded model-based reinforcement learning studies when logged actions are correlated with privileged latent state and develops causal corrections for policy learning \citep{gasse2023confounded}. FOHL, although developed for heterogeneous treatment effects rather than sequential control, provides a relevant factual-abduction principle: infer latent heterogeneity from factual treatment--outcome data before replacing the treatment \citep{zou2023fohl}. Counterfactual Quotient Models directly isolate the part of a future that changes across candidate actions \citep{chen2026cqm}. These approaches clarify how alternative actions should be compared, but they either assume a known or sufficiently accurate structural model, focus on static treatment outcomes, or represent action-dependent outcomes without an object-level causal mechanism. They do not explain which contacts appear, which pathways disappear, and how force propagation changes when a manipulation action is replaced. Their counterfactuals may therefore remain brittle under novel object compositions and mechanism shifts.

We address the intersection of these gaps with \method{}, a counterfactual world model built on an action-grounded causal graph. The latent physical state $P_t$ is abducted from factual interaction history and the action acts as a \emph{graph generator}. For each candidate action, the model constructs a sparse graph $G_t$ and edge weights $W_t$ over the current robot and object set. $G_t$ states which causal pathways are active; $W_t$ states their signed strengths. Shared type-conditioned functions instantiate this graph for any number of objects, and a nested factor graph specifies which state dimensions are transmitted along each active object edge. One physical situation, consisting of latent physics and transition noise, is passed through the graph of every candidate action, so the model represents all candidate futures jointly: replacing $A_t$ regenerates $(G_t,W_t)$ and rolls the resulting mechanisms forward under the same situation. Counterfactual predictions are anchored on the observed successor, and the graph confines each action contrast to the objects it reaches. On this model we build Counterfactual Advantage Planning (CAP). Candidate plans are ranked by their advantage over an incumbent plan within shared situations, switches with a high probability of harm are rejected, and each executed transition is converted into hindsight advantages that update the physics posterior, the incumbent of the next attempt, and an amortized proposal policy.

Consider a robot arm facing two blocks. In the factual trajectory, the gripper pushes the red block, which then collides with the blue block. From the observed displacement and contact response, the model infers latent mass, friction, and compliance. The factual action activates gripper$\rightarrow$red and red$\rightarrow$blue edges. To evaluate a counterfactual grasp, the model retains the same latent physical state but regenerates the graph: the pushing edge is replaced by grasp contacts and the collision edge may disappear. To evaluate a stronger push in another direction, the same edge types may remain active while their weights and downstream collision topology change. A planner can compare these complete trajectories, including success, collision, and uncertainty, rather than attempt to fit a scalar counterfactual value that was never observed. If the factual push fails to separate the blocks, the hindsight advantage identifies the push that would have separated them under the same latent physics and contact noise. That push becomes the incumbent of the retry, which reuses the inferred mass and friction, and a stronger alternative is adopted only if it rarely does worse than the incumbent in the same situations.

Our contributions are:
\begin{itemize}[leftmargin=1.4em,itemsep=0.2em,topsep=0.4em]
    \item We formulate a counterfactual world model as the joint law of all candidate futures, generated by passing one abducted physical situation through an action-grounded causal graph. We show that factual data determine only the marginals of this law, that anchoring with the average effect incurs an error equal to the conditional variance of the individual effect, and we enforce action-invariant abduction with a round-trip loss that needs no counterfactual labels.
    \item We introduce a variable-cardinality object--factor graph in which actions control both edge activation and edge strength, enabling topology-changing interventions such as contact creation, removal, and force redirection.
    \item We propose Counterfactual Advantage Planning, which ranks plans by paired advantages within shared situations, constrains the individual-level probability of harm, and converts executed transitions into hindsight advantages for physics updates, retries, and critic-free advantage-weighted policy learning. Paired advantages reduce estimator variance by the factor $1-\rho$ and cancel model errors common to both branches.
    \item We derive sufficient conditions for $p(S_{t+1}\mid H_t,A_t)=p(S_{t+1}\mid H_t,\doop(A_t))$ and distinguish partial observability from genuine hidden action confounding.
    \item We develop a paired-intervention evaluation that measures counterfactual accuracy, graph recovery, compositional generalization, and decision quality against simulator ground truth.
\end{itemize}

\section{Related Work}

\subsection{Causal World Models}

Structural causal models distinguish conditioning from intervention and counterfactual queries \citep{pearl2009causality}. Causal representation learning extends this distinction to latent variables, seeking independent and reusable mechanisms rather than entangled predictive features \citep{scholkopf2021causalrep}. Causal World Models apply these principles to physical scenes by inferring unobserved object properties and predicting intervention outcomes \citep{li2020cwm}. CausalVAE as a plug-in estimates a sparse latent adjacency matrix and imposes the differentiable acyclicity constraint
\begin{equation}
    h(B)=\operatorname{tr}\!\left(\exp(B\odot B)\right)-d=0,
\end{equation}
following continuous DAG learning \citep{zheng2018notears,ding2026causalvae}. This is suitable for directed relations among a fixed set of same-slice latent variables, but its adjacency parameters do not directly transfer when objects enter or leave a scene.

A second line emphasizes physical context and object compositionality. DALI infers latent environment dynamics from interaction history \citep{roder2025dali}. Factored world models and FOCUS decompose scenes into objects and exploit shared relational dynamics for manipulation and exploration \citep{biza2022factored,ferraro2023focus}. FIOC-WM learns modular object interactions, Dyn-O separates dynamics-aware object factors, and SPARTAN learns sparse context-dependent local structures robust to distractors and dynamics changes \citep{feng2025fioc,wang2025dyno,lei2025spartan}. These models improve transfer, but latent physical context is generally consumed as a global predictor input, while local edges are learned for factual prediction. They do not jointly require that a replacement action regenerate an executable graph under the same abducted physical realization. Our method treats the local graph itself as an action-dependent causal mechanism and separates edge existence from edge strength.

\subsection{Counterfactual World Models and Control}

CF-GPS introduced counterfactual trajectory generation for reinforcement learning by abducting exogenous randomness from an observed trajectory and replacing the policy \citep{buesing2019cfgps}. This gives lower-variance comparisons because factual and counterfactual trajectories share the inferred scenario, but it relies on a correct causal simulator. Confounded model-based RL highlights a complementary problem: if a behavior policy observes privileged state that is omitted from the learner's history, observational and interventional action effects can differ \citep{gasse2023confounded}. FOHL infers latent heterogeneity from factual observations before treatment replacement \citep{zou2023fohl}; this motivates our physical-state abduction, although sequential manipulation requires time-varying objects, actions, and mechanisms rather than a static treatment--outcome graph.

Counterfactual Quotient Models isolate action-dependent future components by comparing candidate predictions and cancelling their common mode \citep{chen2026cqm}. This is useful for control because action-independent prediction error need not corrupt action ranking. However, an action-effect representation does not identify which object interactions produced the effect and offers no direct mechanism for transfer across changing object sets. Our approach generates structured counterfactual trajectories from an action-conditioned graph and evaluates them with model-predictive control. It therefore complements common-mode cancellation while retaining an inspectable and reusable dynamics mechanism. Model-based agents such as DreamerV3 and TD-MPC2 score imagined rollouts with learned critics and sample the future of each candidate independently \citep{hafner2023dreamerv3,hansen2024tdmpc2}; CAP instead computes advantages from coupled counterfactual rollouts that share an abducted situation, which also exposes the individual-level probability of harm.

\section{Method}

\subsection{Causal Model}

Let $H_t=(O_{0:t},A_{0:t-1})$ be the complete recorded history before the current action, $A_t$ the action, and $S_{t+1}$ the task-relevant successor state. $P_t$ denotes latent physical state, while $G_t$ and $W_t$ denote the action-grounded causal graph and its edge strengths. The assumed causal structure is shown first because it determines both the identification result and the architecture.

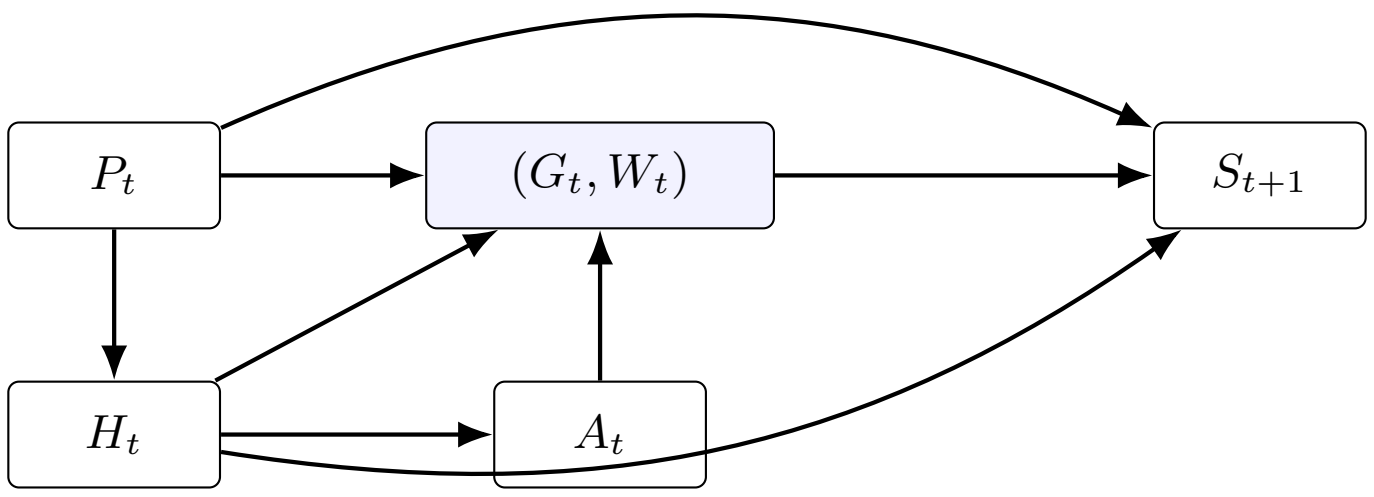
\begin{figure}[t]
\centering
\begin{tikzpicture}[
    >=Latex,
    node distance=10mm and 18mm,
    every node/.style={font=\small},
    var/.style={draw,rounded corners=2pt,minimum width=14mm,minimum height=7mm,fill=white},
    mech/.style={draw,rounded corners=2pt,minimum width=23mm,minimum height=7mm,fill=blue!5},
    line/.style={->,thick}
]
    \node[var] (P) {$P_t$};
    \node[var,below=of P] (H) {$H_t$};
    \node[var,right=of H] (A) {$A_t$};
    \node[mech,above=of A] (GW) {$(G_t,W_t)$};
    \node[var,right=25mm of GW] (S) {$S_{t+1}$};

    \draw[line] (P) -- (H);
    \draw[line] (P) -- (GW);
    \draw[line] (H) -- (A);
    \draw[line] (H) -- (GW);
    \draw[line] (A) -- (GW);
    \draw[line] (GW) -- (S);
    \draw[line] (P) to[bend left=24] (S);
    \draw[line] (H) to[bend right=22] (S);
\end{tikzpicture}
\caption{Two-slice causal model. The behavior policy selects $A_t$ from the recorded history $H_t$. Latent physics $P_t$ affects the history, the action-grounded mechanism $(G_t,W_t)$, and the successor state, but has no direct edge to $A_t$ after conditioning on $H_t$. The current action changes the successor through graph topology and edge strength.}
\label{fig:scm}
\end{figure}

The corresponding structural equations are
\begin{align}
P_t &= f_P(P_{t-1},\epsilon_t^P),
& H_t &= f_H(O_{0:t},A_{0:t-1}), \nonumber\\
A_t &= f_\pi(H_t,\epsilon_t^A),
& (G_t,W_t) &= f_G(P_t,H_t,A_t), \label{eq:scm}\\
S_{t+1} &= f_S(H_t,P_t,G_t,W_t,U_t), \nonumber
\end{align}
where $U_t$ is transition-specific exogenous noise. The graph does not forbid $H_t$ and $A_t$ from being strongly correlated; that dependence is the policy. It asserts only that all variables used by the behavior policy are contained in $H_t$.

The physical state is hierarchical,
\begin{equation}
P_t=\left(P^g_t,\,P^o_t,\,P^r_t\right),\qquad
P^g_t\in\mathbb{R}^{d_g},\quad
P^o_t\in\mathbb{R}^{N_t\times d_o},\quad
P^r_t\in\mathbb{R}^{N_t\times N_t\times d_r},
\label{eq:physical-state}
\end{equation}
where $P^g_t$ captures scene-level physics, the rows of $P^o_t$ hold object properties such as mass and inertia, and $P^r_t$ holds relation properties such as friction or compliance for the candidate pairs $\Cand_t$, with all other entries masked. The object set $\ObjSet_t$, and hence $N_t=|\ObjSet_t|$, may change across scenes and time.

\subsection{From Observational to Interventional Dynamics}

We assume consistency, positivity, modularity of the action intervention, and no hidden action confounding:
\begin{equation}
S_{t+1}(a)\indep A_t\mid H_t.
\label{eq:exchangeability}
\end{equation}
The last condition holds in \cref{fig:scm} because the only incoming arrow to $A_t$ is from $H_t$. It does not require $P_t$ to be observed.

\begin{proposition}[Identification of the one-step action effect]
Under the assumptions above, for every supported $(h,a)$,
\begin{equation}
\boxed{p(s_{t+1}\mid h,a)=p(s_{t+1}\mid h,\doop(a)).}
\label{eq:identification}
\end{equation}
\end{proposition}

\begin{proof}
Using the mechanism variables explicitly, intervention gives
\begin{align}
p(s'\mid h,\doop(a))
&=\int p(s'\mid h,p,g,w)
        p(g,w\mid h,p,\doop(a))
        p(p\mid h,\doop(a))\,d(p,g,w) \nonumber\\
&=\int p(s'\mid h,p,g,w)
        p(g,w\mid h,p,a)
        p(p\mid h)\,d(p,g,w),
\label{eq:do-transition}
\end{align}
where modularity replaces only the action equation. The observational transition is
\begin{equation}
p(s'\mid h,a)=\int p(s'\mid h,p,g,w)
p(g,w\mid h,p,a)p(p\mid h,a)\,d(p,g,w).
\label{eq:obs-transition}
\end{equation}
The graph entails $A_t\indep P_t\mid H_t$, so $p(p\mid h,a)=p(p\mid h)$. Substitution into \cref{eq:obs-transition} yields \cref{eq:identification}. Equivalently, conditioning on $H_t$ blocks the backdoor paths $A_t\leftarrow H_t\leftarrow P_t\rightarrow S_{t+1}$ and $A_t\leftarrow H_t\rightarrow S_{t+1}$.
\end{proof}

The result separates two issues that are often conflated. Partial observability makes $P_t$ uncertain and motivates latent-state inference, but it does not create confounding unless unrecorded information also enters the behavior policy. If a privileged variable $C_t^\pi$ causes both $A_t$ and $S_{t+1}$ after conditioning on $H_t$, then $p(P_t,C_t^\pi\mid h,a)\neq p(P_t,C_t^\pi\mid h)$ and \cref{eq:identification} fails. Randomized exploration, logged privileged inputs, proxy-variable assumptions, or an explicit confounded model are then required \citep{gasse2023confounded}. Positivity is equally essential: a deterministic behavior policy supplies no evidence for unsupported alternatives even when the causal equality is valid.

\subsection{Factual Physical-State Abduction}

The model learns a filter for planning and a smoother for unit-level counterfactuals:
\begin{align}
q_f(P_t\mid H_t), \qquad
q_s(P_t,U_t\mid H_t,A_t,S_{t+1}).
\label{eq:posteriors}
\end{align}
The smoother explains the realized transition by inferring latent physics and exogenous noise. Its posterior follows the hierarchy in \cref{eq:physical-state}:
\begin{equation}
q_s(P_t\mid\tau_t^F)
=q(P_t^g\mid\tau_t^F)\,
q(P^o_t\mid\tau^F_t)\,
q(P^r_t\mid\tau^F_t),
\label{eq:hierarchical-posterior}
\end{equation}
where the object and relation factors are mean-field over the rows of $P^o_t$ and the candidate entries of $P^r_t$, respectively, and are produced by shared amortized encoders.
An amortized filter is trained from the smoother through
\begin{equation}
\mathcal{L}_{\mathrm{dist}}
=\KL\!\left[\operatorname{sg}\{q_s(P_t\mid H_t,A_t,S_{t+1})\}
\,\|\,q_f(P_t\mid H_t)\right],
\label{eq:distillation}
\end{equation}
so prospective control does not use future observations. Episode-persistent priors prevent physical latents from absorbing transient appearance noise.

\subsection{Action-Grounded Object--Factor Graph}

An object encoder produces the node matrix $V_t\in\mathbb{R}^{N_t\times d_v}$, one row per robot part or object in $\ObjSet_t$, and a soft type matrix $\Upsilon_t\in[0,1]^{N_t\times T}$. Types encode interaction roles such as gripper, rigid object, support, or articulated part, rather than instance identity. A shared type template $B^{\mathrm{type}}\in[0,1]^{T\times T}$ and a geometric feasibility mask define the candidate adjacency
\begin{equation}
\bar B_t=\left(\Upsilon_t B^{\mathrm{type}}\Upsilon_t^{\top}\right)\odot K(V_t,A_t)\in[0,1]^{N_t\times N_t}.
\label{eq:candidate}
\end{equation}
The action-conditioned topology and weight matrices are
\begin{align}
G^{\mathrm{obj}}_t
&=\bar B_t\odot
\operatorname{HardConcrete}\!\left(
L_\psi(V_t,P_t,A_t)\right), \label{eq:gate}\\
W^{\mathrm{obj}}_t
&=G^{\mathrm{obj}}_t\odot
\Omega_\eta(V_t,P_t,A_t), \label{eq:weight}
\end{align}
where $L_\psi,\Omega_\eta\in\mathbb{R}^{N_t\times N_t}$ apply shared pairwise networks to every ordered pair of rows of $V_t$. $G^{\mathrm{obj}}_t$ answers which pathways the action activates; $W^{\mathrm{obj}}_t$ gives their signed magnitudes. Hard-concrete gates support differentiable $L_0$ sparsity \citep{louizos2018l0}. Because $L_\psi$ and $\Omega_\eta$ are shared across instances, object addition changes the number of evaluations rather than the number of learned parameters.

Each active object edge is refined over state factors. Splitting every object state into $K$ factors gives the factor matrix $Z_t\in\mathbb{R}^{N_tK\times d_z}$. A Kronecker product lifts the object graph to factor resolution:
\begin{align}
G^{\mathrm{dim}}_t
&=\left(G^{\mathrm{obj}}_t\otimes B^{\mathrm{dim}}\right)\odot
\operatorname{HardConcrete}\!\left(
L_{\psi_d}(Z_t,P_t,A_t)\right), \label{eq:factor-gate}\\
\Gamma_t
&=G^{\mathrm{dim}}_t\odot\left(W^{\mathrm{obj}}_t\otimes\mathbf{1}_K\mathbf{1}_K^{\top}\right), \nonumber\\
\hat Z_{t+1}
&=F_\theta\!\left(Z_t,\,
\operatorname{Agg}\!\left[\Gamma_t\odot M_\theta(Z_t,P_t,A_t)\right],\,U_t\right).
\label{eq:transition}
\end{align}
Here $B^{\mathrm{dim}}\in\{0,1\}^{K\times K}$ is the admissible factor-to-factor template, $\Gamma_t\in\mathbb{R}^{N_tK\times N_tK}$ is the effective weighted adjacency, $M_\theta$ evaluates type-conditioned message functions on all factor pairs, $\odot$ broadcasts over the message dimension, $\operatorname{Agg}$ sums incoming messages over senders, and $F_\theta$ updates each row of $Z_t$ with parameters selected by its type.
The graph connects variables at $t$ to variables at $t+1$ and is therefore acyclic by temporal construction. Unlike a same-slice latent DAG such as CausalVAE, it does not require a NOTEARS penalty unless instantaneous within-slice edges are introduced. Shared edge functions and symmetric aggregation make the transition permutation equivariant, supporting object permutations and variable cardinality; unseen mechanism types still require adaptation or uncertainty detection.

\subsection{Counterfactual World Model}
\label{sec:cwm}

Let $\xi_t=(P_t,U_t)$ denote the physical situation of a transition, consisting of the latent physics and the exogenous noise. Composing the graph generator with the transition in \cref{eq:scm} gives the action-to-outcome map
\begin{equation}
\Phi(H_t,\xi_t,a)=f_S\!\left(H_t,P_t,f_G(P_t,H_t,a),U_t\right),
\label{eq:phi}
\end{equation}
so the potential successor under action $a$ is $S_{t+1}(a)=\Phi(H_t,\xi_t,a)$ and the factual successor is $S^F_{t+1}=S_{t+1}(a^F_t)$. A conventional world model represents, for each action separately, the marginal law $p(S_{t+1}\mid H_t,\doop(A_t{=}a))$. A counterfactual world model represents the joint law of the potential successors of all candidate actions. \method{} obtains this joint law from a single generative program: one situation is drawn and passed through the action-grounded causal graph of every candidate action,
\begin{equation}
\boxed{\;\xi_t\sim q(\xi_t),\qquad
\mathbf{S}_{t+1}(\mathcal{A})=\left\{\Phi(H_t,\xi_t,a)\right\}_{a\in\mathcal{A}},\;}
\label{eq:cwm}
\end{equation}
where $\mathcal{A}$ is a set of candidate actions. The situation is shared across branches; only the graph $f_G(P_t,H_t,a)$ and the mechanisms downstream of it differ. The law $q$ determines which question the model answers:
\begin{align}
\textbf{Prediction:}\quad & q(\xi_t)=p(P_t\mid H_t,A_t{=}a)\,p(U_t), \nonumber\\
\textbf{Intervention:}\quad & q(\xi_t)=p(P_t\mid H_t)\,p(U_t)\;\approx\;q_f(P_t\mid H_t)\,p(U_t), \label{eq:queries}\\
\textbf{Counterfactual:}\quad & q(\xi_t)=p(\xi_t\mid H_t,a^F_t,S^F_{t+1})\;\approx\;q_s(\xi_t\mid H_t,a^F_t,S^F_{t+1}). \nonumber
\end{align}
Prediction conditions on the logged action. Intervention removes the influence of the behavior policy; by \cref{eq:identification} the two laws coincide when $H_t$ contains every policy input. The counterfactual additionally conditions on the realized successor, so every alternative action is evaluated under the mass, friction, contacts, and disturbance that actually occurred.

\paragraph{Abduction, intervention, prediction.}
For a factual transition $\tau_t^F=(H_t,a_t^F,S_{t+1}^F)$, the counterfactual query is executed as
\begin{align}
\textbf{Abduction:}\quad &\xi_t^F=(P_t^F,U_t^F)\sim
q_s(\xi_t\mid H_t,a_t^F,S_{t+1}^F), \nonumber\\
\textbf{Intervention:}\quad &A_t\leftarrow a_t^{\mathrm{cf}}, \nonumber\\
\textbf{Prediction:}\quad &(G_t^{\mathrm{cf}},W_t^{\mathrm{cf}})
=f_G(P_t^F,H_t,a_t^{\mathrm{cf}}), \label{eq:cf-regeneration}\\
&\hat S_{t+1}^{\mathrm{cf}}
=f_S(H_t,P_t^F,G_t^{\mathrm{cf}},W_t^{\mathrm{cf}},U_t^F). \nonumber
\end{align}
Holding $\xi_t^F$ fixed makes the comparison unit-level: only the action equation and its downstream graph change. For multi-step counterfactuals, the predicted object set, candidate relations, gates, and weights are recomputed at every step. Thus a new contact may activate an edge, an intervention may sever a parent relation, and subsequent propagation is automatically pruned.

\paragraph{Anchored counterfactual prediction.}
The counterfactual successor is written as the observed successor plus an action contrast:
\begin{equation}
\boxed{\hat S^{\mathrm{cf}}_{t+1}(a')=S^F_{t+1}+\hat\Delta_t(a',a^F_t),\qquad
\hat\Delta_t(a',a^F_t)=\Phi(H_t,\xi^F_t,a')-\Phi(H_t,\xi^F_t,a^F_t).}
\label{eq:anchor}
\end{equation}
Under an exact smoother, $\Phi(H_t,\xi^F_t,a^F_t)=S^F_{t+1}$ and \cref{eq:anchor} coincides with \cref{eq:cf-regeneration}. Under an approximate smoother, errors common to both actions cancel, and the model is responsible only for the change caused by the replacement action. The graph localizes this change. Let $\mathbf{n}_t=(G^{\mathrm{obj},F}_t+G^{\mathrm{obj},\mathrm{cf}}_t)^{\top}\mathbf{1}_{N_t}$ be the in-degree of each object under the factual and counterfactual graphs. An object with no active incoming edge under either action receives zero messages in both branches and shares $Z_t$ and $U_t$, so
\begin{equation}
\operatorname{diag}\!\left(\mathbb{I}[\mathbf{n}_t=\mathbf{0}]\right)\hat\Delta_t(a',a^F_t)=\mathbf{0},
\label{eq:locality}
\end{equation}
where $\hat\Delta_t$ is arranged with one row per object. Objects outside the regenerated mechanism therefore keep their observed successors exactly.

\paragraph{What factual data determine.}
Let $\Delta_t(a',a)=S_{t+1}(a')-S_{t+1}(a)$ be the individual action effect and $\mu(h,a)=\E[S_{t+1}\mid H_t=h,A_t=a]$ the factual conditional mean. \Cref{eq:identification} identifies the average effect,
\begin{equation}
\E\!\left[\Delta_t(a',a^F_t)\mid H_t\right]=\mu(H_t,a')-\mu(H_t,a^F_t),
\label{eq:average-effect}
\end{equation}
and anchoring with this average contrast in place of $\hat\Delta_t$ incurs
\begin{equation}
\E\!\left[\big\|S^F_{t+1}+\mu(H_t,a')-\mu(H_t,a^F_t)-S_{t+1}(a')\big\|^2\,\middle|\,H_t\right]
=\operatorname{tr}\operatorname{Var}\!\left[\Delta_t(a',a^F_t)\mid H_t\right].
\label{eq:mean-anchor-error}
\end{equation}
The individual effect itself lies beyond the reach of factual data.

\begin{proposition}[Factual data determine marginals, not couplings]
\label{prop:coupling}
There exist two structural models with identical interventional laws $p(S_{t+1}\mid H_t,\doop(A_t{=}a))$ for every action $a$, and hence identical factual likelihoods and identical average effects, whose individual effects $\Delta_t(a',a)$ have different laws.
\end{proposition}

\begin{proof}
Let $U_t\sim\mathcal{N}(\mathbf{0},I)$. Model~1 sets $S_{t+1}(a)=S_{t+1}(a')=U_t$; Model~2 sets $S_{t+1}(a)=U_t$ and $S_{t+1}(a')=-U_t$. Every marginal is $\mathcal{N}(\mathbf{0},I)$, whereas $\Delta_t(a',a)=\mathbf{0}$ under Model~1 and $\Delta_t(a',a)=-2U_t$ under Model~2.
\end{proof}

Every quantity that depends on the individual effect, including the error in \cref{eq:mean-anchor-error}, the regret of an executed action, and the probability that switching actions causes harm, is therefore fixed by the coupling in \cref{eq:cwm} rather than by factual fit \citep{pearl2009causality,tian2000probabilities}. \method{} determines the coupling structurally: the action enters only through the graph generator, and the situation $\xi_t$ is shared by all branches. Manipulation is where this matters. Whether a push transfers motion to a second block depends on mass and friction, so the same action creates a contact edge in one situation and not in another, and $\operatorname{tr}\operatorname{Var}[\Delta_t\mid H_t]$ is large. Because $\Delta_t$ is a deterministic function of $(H_t,\xi_t)$, $\operatorname{Var}[\Delta_t\mid H_t,\xi_t]=\mathbf{0}$; conditioning on the abducted situation selects the graph that $a'$ generates in the realized scene, and \cref{eq:anchor} carries the corresponding situation-specific contrast.

\subsection{Learning}

The model is trained end to end with
\begin{equation}
\mathcal{L}
=\mathcal{L}_{\mathrm{dyn}}
+\beta\mathcal{L}_{\mathrm{KL}}
+\lambda_s\mathcal{L}_{\mathrm{sparse}}
+\lambda_p\mathcal{L}_{\mathrm{persist}}
+\lambda_d\mathcal{L}_{\mathrm{dist}}
+\lambda_{rt}\mathcal{L}_{\mathrm{rt}}
+\lambda_{cf}\mathcal{L}_{\mathrm{cf}}.
\label{eq:loss}
\end{equation}
$\mathcal{L}_{\mathrm{dyn}}$ is a multi-step state and observation likelihood; $\mathcal{L}_{\mathrm{KL}}$ regularizes latent physics and noise; $\mathcal{L}_{\mathrm{sparse}}=\|G^{\mathrm{obj}}_t\|_1+\alpha\|G^{\mathrm{dim}}_t\|_1$ selects compact mechanisms; and $\mathcal{L}_{\mathrm{persist}}$ enforces temporal consistency of physical properties.

Factual likelihoods constrain the marginals of \cref{eq:cwm}. The coupling additionally requires that the abducted situation not depend on which action was taken. The round-trip loss enforces this without counterfactual labels: an alternative action is sampled, its anchored counterfactual is generated, the situation is abducted back from that counterfactual, and the factual successor must be recovered under the factual action,
\begin{equation}
\mathcal{L}_{\mathrm{rt}}=\E_{a'\sim\rho(\cdot\mid H_t)}\;\E_{\tilde\xi_t\sim q_s(\cdot\mid H_t,a',\hat S^{\mathrm{cf}}_{t+1}(a'))}\;
d\!\left(\Phi(H_t,\tilde\xi_t,a^F_t),\,S^F_{t+1}\right),
\label{eq:roundtrip}
\end{equation}
where $\rho$ is the action proposal. A smoother that stores action-specific information in $\xi_t$, or a mechanism that routes situation effects through the action, fails the round trip. When a simulator can clone a factual state and execute another action under a shared random seed,
\begin{equation}
\mathcal{L}_{\mathrm{cf}}
=d(\hat S_{t+1}^{\mathrm{cf}},S_{t+1}^{\mathrm{cf}})
+\lambda_g d_G(G_t^{\mathrm{cf}},G_t^{\mathrm{cf},*})
\label{eq:cf-loss}
\end{equation}
provides paired supervision. Without graph labels, $d_G$ is omitted and graph validity is tested by held-out interventions, object removal, and invariance to non-parent distractors.

\subsection{Counterfactual Advantage Planning}
\label{sec:cap}

A decision is a comparison between actions. Counterfactual Advantage Planning (CAP) makes this comparison inside shared physical situations, which a world model with only marginal laws cannot do.

\paragraph{Counterfactual advantage.}
For an action sequence $\mathbf{a}=a_{t:t+H-1}$ and a situation $\xi=(P_t,U_{t:t+H-1})$, let $\Phi_{t:t+H}(H_t,\xi,\mathbf{a})$ be the rollout obtained by iterating \cref{eq:phi} with the graph regenerated at every step, and let $J(\xi,\mathbf{a})=\sum_{h=0}^{H-1}r(\hat S_{t+h},a_{t+h})$ be its return. The counterfactual advantage of $\mathbf{a}$ over a reference sequence $\bar{\mathbf{a}}$ is
\begin{equation}
\boxed{\;A(\mathbf{a};\bar{\mathbf{a}},\xi)=J(\xi,\mathbf{a})-J(\xi,\bar{\mathbf{a}}),\qquad \xi\sim q(\xi).\;}
\label{eq:cf-advantage}
\end{equation}
Before acting, $\bar{\mathbf{a}}$ is the incumbent plan, namely the previous plan shifted by one step, and $q=q_f(P_t\mid H_t)\,p(U)$. In hindsight, $\bar{\mathbf{a}}=\mathbf{a}^F$, $q=q_s(\cdot\mid\tau^F)$ for the factual segment $\tau^F=(H_t,\mathbf{a}^F,\mathbf{S}^F)$, and both branches are anchored on $\mathbf{S}^F$ as in \cref{eq:anchor}. The mean $\E_\xi[A]=V(\mathbf{a})-V(\bar{\mathbf{a}})$ is the interventional advantage and is fixed by the marginals. The distribution of $A$ over $\xi$ is the individual advantage and exists only in a counterfactual world model. CAP uses both.

\paragraph{Paired estimation.}
With $M$ shared draws $\xi^{(1)},\dots,\xi^{(M)}$, the estimator $\hat A_M=\frac{1}{M}\sum_{m}A(\mathbf{a};\bar{\mathbf{a}},\xi^{(m)})$ evaluates both sequences in the same situations.

\begin{proposition}[Variance of paired advantages]
\label{prop:paired}
Let $\sigma^2_{\mathbf{a}}=\operatorname{Var}_\xi J(\xi,\mathbf{a})$, $\sigma^2_{\bar{\mathbf{a}}}=\operatorname{Var}_\xi J(\xi,\bar{\mathbf{a}})$, and $\rho_{\mathbf{a}\bar{\mathbf{a}}}=\operatorname{Corr}_\xi\!\left[J(\xi,\mathbf{a}),J(\xi,\bar{\mathbf{a}})\right]$. Then
\begin{equation}
\operatorname{Var}\!\left[\hat A_M\right]=\frac{\sigma^2_{\mathbf{a}}+\sigma^2_{\bar{\mathbf{a}}}-2\rho_{\mathbf{a}\bar{\mathbf{a}}}\,\sigma_{\mathbf{a}}\sigma_{\bar{\mathbf{a}}}}{M},
\label{eq:paired-var}
\end{equation}
whereas evaluating the two sequences on independent draws gives $(\sigma^2_{\mathbf{a}}+\sigma^2_{\bar{\mathbf{a}}})/M$. For $\sigma_{\mathbf{a}}=\sigma_{\bar{\mathbf{a}}}$, pairing reduces the number of rollouts required for a given ranking accuracy by the factor $1-\rho_{\mathbf{a}\bar{\mathbf{a}}}$.
\end{proposition}

\begin{proof}
The summands are i.i.d.\ with variance $\operatorname{Var}_\xi[J(\xi,\mathbf{a})-J(\xi,\bar{\mathbf{a}})]=\sigma^2_{\mathbf{a}}+\sigma^2_{\bar{\mathbf{a}}}-2\rho_{\mathbf{a}\bar{\mathbf{a}}}\sigma_{\mathbf{a}}\sigma_{\bar{\mathbf{a}}}$; with independent draws the covariance term is zero. Under the normal approximation, the probability of ranking $\mathbf{a}$ and $\bar{\mathbf{a}}$ incorrectly is $\tfrac12\operatorname{erfc}\!\big(|V(\mathbf{a})-V(\bar{\mathbf{a}})|/(2\operatorname{Var}[\hat A_M])^{1/2}\big)$, which depends on $M$ only through $M/\operatorname{Var}[\hat A_1]$.
\end{proof}

In manipulation, most return variance comes from the shared situation, namely mass, friction, object poses, and disturbances, so $\rho_{\mathbf{a}\bar{\mathbf{a}}}$ approaches one for nearby action sequences. These are exactly the candidates that CEM and MPPI must discriminate in their final iterations.

\paragraph{Common-mode cancellation.}
Write the model return as $J_\theta(\xi,\mathbf{a})=J(\xi,\mathbf{a})+e_0(\xi)+e_1(\xi,\mathbf{a})$, where $e_0$ collects errors that do not depend on the action. The advantage error $e_1(\xi,\mathbf{a})-e_1(\xi,\bar{\mathbf{a}})$ does not contain $e_0$, and ranking by $\E_\xi[A_\theta]$ is correct whenever $|V(\mathbf{a})-V(\bar{\mathbf{a}})|>|\E_\xi[e_1(\xi,\mathbf{a})-e_1(\xi,\bar{\mathbf{a}})]|$. The causal graph determines what is common. The objects whose rows can differ between the two branches are tracked by
\begin{equation}
\mathbf{d}_{h+1}=\mathbb{I}\!\left[\mathbf{d}_h+\big(G^{\mathrm{obj}}_{t+h}(\mathbf{a})+G^{\mathrm{obj}}_{t+h}(\bar{\mathbf{a}})\big)^{\top}\mathbf{1}_{N_t}>\mathbf{0}\right],\qquad \mathbf{d}_0=\mathbf{0},
\label{eq:support}
\end{equation}
which extends \cref{eq:locality} over the horizon. Rows with $\mathbf{d}_h=0$ are identical in both branches and are copied from the reference rollout; for object-additive rewards, their errors enter $e_0$ and cancel. Messages and updates are recomputed only for flagged rows, so the cost of scoring a candidate scales with the part of the scene that its action reaches.

\paragraph{Harm-constrained decision.}
The individual advantage gives the probability that switching from the incumbent hurts in the same situation,
\begin{equation}
\pi_{\mathrm{harm}}(\mathbf{a};\bar{\mathbf{a}})=\Pr_{\xi}\!\left[A(\mathbf{a};\bar{\mathbf{a}},\xi)<-\delta\right].
\label{eq:harm}
\end{equation}
By \cref{prop:coupling}, $\pi_{\mathrm{harm}}$ is not a function of the marginal laws; computing it from independent draws replaces the coupling by independence and misstates harm in both directions. CAP selects
\begin{equation}
\mathbf{a}^{*}=\arg\max_{\mathbf{a}}\;\E_{\xi}\!\left[A(\mathbf{a};\bar{\mathbf{a}},\xi)\right]-\lambda_u\,\mathcal{U}(\mathbf{a})
\quad\text{s.t.}\quad \pi_{\mathrm{harm}}(\mathbf{a};\bar{\mathbf{a}})\le\epsilon,
\qquad \xi\sim q_f(P_t\mid H_t)\,p(U),
\label{eq:planning}
\end{equation}
where $\mathcal{U}$ is ensemble disagreement over rollouts. CEM or MPPI scores elites with the Lagrangian of \cref{eq:planning}, estimating both terms from the same $M$ shared draws. The incumbent has $A\equiv 0$ and $\pi_{\mathrm{harm}}=0$, so it is always feasible: the controller changes its plan only when the improvement holds within shared situations. It executes the first action of $\mathbf{a}^{*}$ and replans, and it requires no targets for the unobserved quantity $Q(H_t,a^{\mathrm{cf}})$.

\paragraph{Hindsight advantage.}
After the factual segment $\tau^F$ is observed, the same advantage is evaluated in the abducted situation with the executed sequence as reference:
\begin{equation}
\mathbf{a}^{\mathrm{cf}}=\arg\max_{\mathbf{a}}\;\E_{\xi\sim q_s(\cdot\mid\tau^F)}\!\left[A(\mathbf{a};\mathbf{a}^F,\xi)\right],\qquad
\mathcal{R}(\tau^F)=\E_{\xi\sim q_s(\cdot\mid\tau^F)}\!\left[A(\mathbf{a}^{\mathrm{cf}};\mathbf{a}^F,\xi)\right]\ge 0.
\label{eq:retrospective}
\end{equation}
The regret is nonnegative because $A(\mathbf{a}^F;\mathbf{a}^F,\xi)\equiv 0$. It separates failures caused by the action choice ($\mathcal{R}>0$) from failures that no action could have avoided in that physical situation ($\mathcal{R}=0$). The hindsight advantage drives three updates. First, a retry keeps the episode-persistent physics and resamples the transient noise, and it starts from the hindsight-optimal plan as its incumbent:
\begin{equation}
q_{\mathrm{retry}}(\xi)=q_s(P\mid\tau^F)\,p(U),\qquad \bar{\mathbf{a}}_{\mathrm{retry}}=\mathbf{a}^{\mathrm{cf}}.
\label{eq:retry}
\end{equation}
Mass, friction, and compliance recur in the next attempt while the previous disturbance does not, so the retry is planned with \cref{eq:planning} under $q_{\mathrm{retry}}$. Second, a proposal policy $\pi_\phi$ that initializes CEM and MPPI is trained by counterfactual advantage-weighted regression,
\begin{equation}
\mathcal{L}_{\pi}=-\E_{\tau^F}\;\E_{\mathbf{a}\sim\rho(\cdot\mid H_t)}\!\left[\exp\!\left(\bar A(\mathbf{a};\mathbf{a}^F)/\eta\right)\log\pi_\phi(\mathbf{a}\mid H_t)\right],
\qquad \bar A(\mathbf{a};\mathbf{a}^F)=\E_{\xi\sim q_s(\cdot\mid\tau^F)}\!\left[A(\mathbf{a};\mathbf{a}^F,\xi)\right],
\label{eq:cawr}
\end{equation}
which has the form of advantage-weighted regression \citep{peng2019awr} with the advantage computed from coupled counterfactual rollouts in the abducted situation instead of a learned critic. Every logged transition thus supplies advantages for many actions that were never executed. Third, candidate actions whose counterfactual rollouts collide under the abducted physics are added to the safety screen of subsequent decisions.

CAP closes the decision loop: prospective advantages choose the action, the factual outcome updates the situation posterior, and hindsight advantages turn the same outcome into supervision for the next decision.

\section{Experimental Design}

\paragraph{Environments and data.}
CausalWorld supplies controlled interventions on masses, sizes, colors, robot properties, and task variables \citep{ahmed2020causalworld}. Multi-task robotic suites such as Meta-World or ManiSkill-style environments add pushing, grasping, placement, articulation, and variable object composition. At selected factual states, the simulator is cloned and alternative actions are executed with matched initial state and random seed. This produces paired unit-level counterfactual trajectories and, when simulator contacts are exposed, graph supervision.

\paragraph{Generalization protocol.}
Training and test distributions are separated along object instance, object count, layout, mechanism composition, mass and friction, policy, and distractor dimensions. Interpolation and extrapolation of physical parameters are reported separately. Factual-only training and training with paired counterfactual supervision are also separated so that intervention data requirements remain explicit.

\paragraph{Baselines.}
Predictive baselines include recurrent latent world models, TD-MPC-style models, dense Transformers, and fully connected GNNs \citep{hafner2023dreamerv3,hansen2024tdmpc2}. Causal and object-centric baselines include Causal World Models, CausalVAE plug-in, DALI, factored world models, FOCUS, FIOC-WM, Dyn-O, and SPARTAN \citep{li2020cwm,ding2026causalvae,roder2025dali,biza2022factored,ferraro2023focus,feng2025fioc,wang2025dyno,lei2025spartan}. Counterfactual baselines include CF-GPS-style shared-noise rollouts, independent candidate rollouts, and Counterfactual Quotient Models \citep{buesing2019cfgps,chen2026cqm}. Decision baselines include MPC on expected return with independent draws, critic-based advantages from TD-MPC2-style value functions, and advantage-weighted regression with a learned critic \citep{hansen2024tdmpc2,peng2019awr}.

\paragraph{Metrics and ablations.}
We measure one-step and long-horizon factual error, paired counterfactual state and action-effect error, action-ranking regret, contact prediction, graph precision/recall and structural Hamming distance, calibration, and control success, collision rate, and sample efficiency. Decisions are scored against the cloned simulator. Prospective quality is measured by ranking accuracy as a function of the number of rollouts $M$, which tests \cref{prop:paired}, and by the realized harm rate, the fraction of plan switches that lower the return relative to the incumbent under the same seed, compared with the target $\epsilon$ in \cref{eq:planning}. Hindsight quality is measured by the fraction of failed episodes in which $\mathbf{a}^{\mathrm{cf}}$ from \cref{eq:retrospective} reaches the goal under the factual seed, by the agreement between the estimated regret $\mathcal{R}$ and the simulator regret, and by retry success under \cref{eq:retry} with a fresh seed. Key ablations remove latent physics, graph regeneration, object- or factor-level gates, edge weights, sparsity, smoother-to-filter distillation, type sharing, the round-trip loss \cref{eq:roundtrip}, and paired counterfactual supervision, and replace the abducted contrast $\hat\Delta_t$ by the average contrast of \cref{eq:average-effect}. Planner ablations replace shared draws by independent draws, drop the harm constraint, drop the hindsight updates, and replace the counterfactual advantage in \cref{eq:cawr} by a learned critic. A privileged-policy condition deliberately violates \cref{eq:exchangeability} to test the identification boundary.



\section{Conclusion}

Conventional world models predict what usually follows an action but need not discover why. Causal world models expose reusable mechanisms, while counterfactual models compare alternative actions in the same factual situation. \method{} combines these requirements in a counterfactual world model: it abducts latent physics, treats the action as a generator of sparse graph topology and edge strength, and passes one shared situation through the regenerated graph of every candidate action. Counterfactual Advantage Planning turns the resulting coupled futures into decisions, comparing plans within shared situations, bounding the probability of harm, and learning from the actions that were not taken. The resulting research question is testable: whether explicit action-grounded mechanisms improve counterfactual accuracy, decision quality, compositional generalization, and robotic control beyond equally expressive predictive, object-centric, and sparse-attention models.

\bibliographystyle{plainnat}
\bibliography{references}

\end{document}